\documentclass{bmvc2k}

\usepackage{microtype}
\usepackage{graphicx}
\usepackage{booktabs} % for professional tables

\usepackage{amsmath,amsfonts,bm}

\def\eqref#1{equation~\ref{#1}}
\def\floor#1{\lfloor #1 \rfloor}
\def\1{\bm{1}}

\def\eps{{\epsilon}}

\DeclareMathAlphabet{\mathsfit}{\encodingdefault}{\sfdefault}{m}{sl}
\SetMathAlphabet{\mathsfit}{bold}{\encodingdefault}{\sfdefault}{bx}{n}

\usepackage{xcolor}
\usepackage{mystyle}
\usepackage{stackengine} 
\usepackage{dblfloatfix} 
\usepackage{pifont}
\usepackage{booktabs}
\usepackage{mathtools}
\usepackage{multirow}
\usepackage{enumitem}

\usepackage{wrapfig}
\usepackage{tabularx}
 
\newcommand\oast{\stackMath\mathbin{\stackinset{c}{0ex}{c}{0ex}{\ast}{\bigcirc}}}
\newcommand{\norm}[1]{\left\lVert#1\right\rVert}

\title{
    Defensive Tensorization
}

\addauthor{Adrian Bulat}{adrian@adrianbulat.com}{{\large*},1}
\addauthor{Jean Kossaifi}{jean.kossaifi@gmail.com}{{\large*},2}
\addauthor{Sourav Bhattacharya}{sourav.b1@samsung.com}{1}
\addauthor{Yannis Panagakis}{yannisp@di.uoa.gr}{3}
\addauthor{Timothy Hospedales}{t.hospedales@ed.ac.uk}{1,4}
\addauthor{Georgios Tzimiropoulos}{g.tzimiropoulos@qmul.ac.uk}{1,5}
\addauthor{Nicholas D Lane}{nic.lane@samsung.com}{1,6}
\addauthor{Maja Pantic}{m.pantic@imperial.ac.uk}{7}

\addinstitution{
 Samsung AI Centre\\
 Cambridge, UK
}
\addinstitution{
 NVIDIA\\
 Santa Clara, USA
}
\addinstitution{
 University of Athens\\
 Athens, Greece
}
\addinstitution{
University of Edinburgh\\
Edinburgh, UK
}
\addinstitution{
 Queen Mary University London\\
 London, UK
}
\addinstitution{
 University of Cambridge\\
 Cambridge, UK
}
\addinstitution{
 Imperial College London\\
 London, UK
}

\runninghead{Bulat and Kossaifi et al.}{Defensive Tensorization}

\begin{document}

\maketitle

\begin{abstract}
    We propose defensive tensorization, an adversarial defence technique that leverages a latent high-order factorization of the network. The layers of a network are first expressed as factorized tensor layers. Tensor dropout is then applied in the latent subspace, therefore resulting in dense reconstructed weights, without the sparsity or perturbations typically induced by the randomization.
    Our approach can be readily integrated with any arbitrary neural architecture and combined with techniques like adversarial training.
    We empirically demonstrate the effectiveness of our approach on standard image classification benchmarks.
    We validate the versatility of our approach across domains and low-precision architectures by considering an audio classification task and binary networks. In all cases, we demonstrate improved performance compared to prior works.
\end{abstract}

\section{Introduction}
The popularity of DNNs in production-ready systems has raised a serious security concern
as DNNs were found to be susceptible to a wide range of adversarial attacks~\citep{madry2017towards,akhtar2018threat,dong2018boosting,huang2017adversarial,goodfellow2014explaining,kurakin2016adversarial}, 
where small and imperceptible perturbations of the input data lead to incorrect predictions by the networks. This shortcoming is an obstacle in wide-scale adoption of DNN, especially when such models become part of security and safety-related solutions~\citep{amodei2016concrete}. 
Consequently, a large volume of work attempts to design robust networks~\citep{dhillon2018stochastic,lin2019defensive,samangouei2018defense, song2017pixeldefend,guo2017countering}. 
However, advances in designing robust DNNs have been followed with stronger perturbation schemes that defeat such defences~\citep{athalye2018obfuscated}.

Most defenses that rely on randomization apply randomized transformations either to the input, e.g.~\citep{xie2018mitigating}, or within the network, e.g., on the activations~\citep{dhillon2018stochastic} or on the weights directly~\citep{wang2018defensive}. However, all these approaches typically introduce artifacts (e.g. sparsity in the weights or activations) and can be defeated by carefully crafted attacks~\citep{athalye2018obfuscated}. 
In this paper, we take a different approach to randomization. We first parametrize the network using tensor factorization, effectively introducing a latent subspace spanning the weights. We then apply randomization in that latent subspace, enabling us to create models that are robust to adversarial attacks, without modifying directly the weights, activations or inputs.
In summary, we make the following contributions:
\begin{itemize}[leftmargin=*]
  \setlength\itemsep{-0.2em}
    \item We propose an adversarial defence technique that relies on a latent randomized tensor parametrization of each layer and can be seamlessly integrated within any architecture.
    \item We thoroughly evaluate our method's robustness against various adversarial attacks and show that it consistently and significantly improves over the current state of the art especially when combined with adversarial training.
    \item We empirically demonstrate that our method successfully hardens the models against these attacks for both quantized and real-valued nets.
    \item We validate our strategy across domains on both image and audio-based classification.

\end{itemize}

\section{Related work}\label{sec:related-work}

\subsection{Adversarial attacks}\label{ssec:related-work-adversarial-attacks}
First, we review a few of the most popular adversarial attacks alongside the current defense strategies employed.
Given a data sample, e.g., an image $\mytensor{X}$, an adversary will try to find a \emph{small} perturbation, often \emph{imperceptible} to a human eye, but that, added to the input sample, will cause it to be misclassified by the target model, with high confidence. Mathematically, the attacker generates a perturbation $\Delta$ bounded in terms of some \(\ell_p\) norm, i.e. $\norm{\Delta}_{p}\leq \epsilon$, typically with \(p = 2\) or \(p = \infty\). The adversarial sample is obtained by adding the perturbation to the input sample $\mytensor{X}_{adv} = \mytensor{X}+\Delta$.

Several ways of obtaining this adversarial perturbations have been proposed~\citep{goodfellow2014explaining,athalye2018obfuscated,carlini2017towards}. Among them, \emph{black-box} attacks, consider the network as a black-box in which the attacker has no information regarding its architecture or the gradients. \emph{White-box} attacks on the contrary, assume complete access to the network architecture and all its parameters. Moreover, attacks can either be \emph{untargeted}, in which case the goal is simply to make the network predict \emph{any} wrong label, or \emph{targeted}, in which case the aim is to force the network to predict a specific label, independently from the input sample. Next, we introduce the main white-box attacks used in this paper.

\textbf{Fast Gradient Sign Method (FGSM)} is a single-step, gradient based technique, introduced by~\citet{goodfellow2014explaining} to generate $\ell_{\infty}$-bounded adversarial perturbations as follows:
\begin{equation}
\mytensor{X}_{adv} = \myvector{X} + \epsilon \cdot \text{sgn}(\nabla_{\mytensor{X}}\mathcal{L}(\theta, \mytensor{X}, y))
\end{equation}
with $\theta$ the parameters of the target network. While the single gradient-step nature of FGSM makes it better for transferability attacks, it can also lead to a suboptimal ascent direction.

\textbf{Basic Iterative Method (BIM) and Projected Gradient Descend (PGD)} aim to address the shortcoming of FGSM by running it for several iterations. \citet{kurakin2016adversarial} propose BIM, in which the FGSM is run for several iterations, clipping the values of the perturbation at each step to be inside the bounds. \citet{madry2017towards} further improve upon this by prepending BIM with a random start and replacing clipping with a projection onto the acceptable set:
\begin{equation}
    \mytensor{X}_{adv}^{t+1} = \Pi_{\mytensor{X}+S}(\mytensor{X}^{t}_{adv}+\alpha \cdot \text{sgn}\left(\nabla_{\mytensor{X}}\mathcal{L}(\theta,X_{adv}^{t},y)\right),
\end{equation}
where $\alpha$ is the step size and $\Pi_{\mytensor{X}+S}$ is a projection operation forcing the generated adversarial samples to be in the $\ell_p$ ball $S$ around $\myvector{X}$. A model resilient to PGD attacks is considered to be reasonably resistant to all first order attacks~\citep{madry2017towards}.

\subsection{Adversarial defences}\label{ssec:related-adversarial-defenses}
Despite recent advances, developing robust neural networks remains an open, challenging  problem~\citep{athalye2018obfuscated}. Current defense strategies typically attempt to either detect the adversarial samples and denoise them, or inject adversarial samples during training. 
The latter is known as adversarial training ~\citep{goodfellow2014explaining,kurakin2016adversarial,madry2017towards} and is considered the most resilient defense technique. As adversarial training degrades the accuracy on the clean data, recent methods attempt to balance-out the trade-off between standard and robust accuracy via boosted loss functions~\citep{wang2020improving}, early stopping ~\citep{rice2020overfitting} or by analyzing a plethora of architectural and training aspects~\citep{gowal2020uncovering}.
%However, the above mentioned defences typically do not increase the robustness to black-box attacks. In addition they can typically be defeated using two-step approaches~\citep{tramer2017ensemble}. 
Beyond this, other defense strategies were proposed. For example, feature squeezing~\citet{xu2017feature} hardens the model by reducing the complexity of the data representation, causing the adversarial perturbations to disappear due to lower sensitivity. \citet{guo2017countering} proposes a set of five transformations, that applied to the image to increase the robustness to adversarial attacks of a given model. 
However, even the combination of all these transformations was shown to be vulnerable to carefully tuned attacks~\citep{athalye2018obfuscated}. \citet{samangouei2018defense} introduces the so-called \emph{Defense-GAN} technique. The main idea is to project the samples into the manifold of a generator before classifying them. Similarly, \citet{song2017pixeldefend} uses a PixelCNN instead of a generative model. \citep{xie2019feature} introduces a series of denoising blocks that perform feature denoising using the non-local means or other filters. \citep{mustafa2019adversarial} proposes to increase the outer-class distance by forcing the features for each class to lie inside a convex polytope that is maximally separated from the ones of other classes. \citep{yang2019me} attempt to scramble the structure of the adversarial noise by randomly dropping pixels out of the input image and then reconstructing them using matrix estimation methods. Despite the variety of recently proposed defence strategies, in \citep{athalye2018obfuscated} the authors show that most of the existing defense techniques rely on one form of gradient obfuscation proposing both a method to detect such class of defences and to defeat them. In parallel, a series of methods propose certified defenses~\citep{zhang2019towards,gowal2019scalable,cohen2019certified}. However, the provided guarantees to not match the empirical result offered by other techniques.

As opposed to all the aforementioned works which either manipulate the data samples~\citep{samangouei2018defense, song2017pixeldefend,guo2017countering} or
introduce stochasticity on the activations~\citep{dhillon2018stochastic} or weights~\cite{wang2018defensive} of each layer, we propose a novel defense strategy that leverages tensor factorization of the weights in order to apply randomization in that latent space, before reconstructing the weights. The approach is introduced in details in section~\ref{sec:method}. 

\subsection{Tensor methods in deep learning and tensor dropout}\label{ssec:related-work-tensors}
Tensors are high dimensional generalizations of matrices~\citep{kolda2009tensor}. Recently, tensor decompositions have found a surge of applications in deep learning, mainly focusing on network compression and acceleration. By parametrizing layers of neural networks using tensor decomposition, or even whole networks~\citep{kossaifi2019t}, the number of parameters can be reduced with little to no loss of performance, and in some cases the operations can be done more efficiently~\citep{lebedev2015speeding,novikov2015tensorizing,yong2016compression,astrid2017cp}.

In some cases, tensor decompositions can exhibit high computational cost and possibly low convergence rates when applied to massive data. To accelerate computation, and enable them to scale, several randomized tensor decompositions have been developed. In this way, CP decomposition can be done by selecting randomly elements from the original tensor~\citep{battaglino2018practical}, or using randomization to solve the problem on one or several smaller tensors before projecting back the result to the original space~\citep{erichson2017randomized,sidiropoulos2014parallel,vervliet2014breaking}. \citet{wang2015fast} proposed a fast yet provable randomized CP decomposition using FFT to perform tensor contraction.
Randomization approaches have also be explored for fast approximation of other tensor decompositions, e.g., Tucker decomposition via sketching~\citep{tsourakakis2010mach, zhou2014decomposition} and tensor ring using tensor random projections~\citep{yuan2019randomized}.

These methods are orthogonal to our approach and can be combined with it. As opposed to the aforementioned works, which focus on compression or efficiency, we explore tensor factorization methods in the context of adversarial defense, proposing to combine a factorized reparametrization of convolutional layers combined with tensor dropout that significantly hardens the model, increasing its robustness to a wide range of adversarial attacks. Our method is generic and can be applied to both real-valued networks and binary ones. In addition, it is orthogonal to the existing defence methods and can be combined with existing defenses such as adversarial training. The method is introduced in details in section~\ref{sec:method}.

\section{Defensive tensorization}\label{sec:method}
Our defense leverages a randomized higher-order factorization method, which is used as the basis for our defense. Typically, defensive methods relying on randomization do so by introducing sparsity in either the weights or the input activation tensors of the layers of the deep neural neural network. For instance, \citet{dhillon2018stochastic} sparsify the input tensor, by stochastically pruning some of the activations and scaling up the remaining ones. \citet{wang2018defensive} apply sparsification to the weights directly using dropout both during training and testing. 

However, all these approaches degrade performance by setting some of the activations to zero, and, while rescaling the non-zero entries can mitigate the issue, increasing the sparsity (and therefore the efficiency of the defense) translates into large losses in performance. By contrast, we propose to rely on a latent parametrization of the layers using tensor decomposition. Intuitively, a latent subspace spanning the weights is learnt, along with a non-linear projection to and from that subspace. The sparsity inducing randomization is applied in the latent space. Upon projection, the resulting weights are dense and yet preserve the robustness against adversarial attacks. This allows us to built models that are both more robust to adversarial methods than existing works, while preserving high classification accuracy.

\textbf{Notation: }
We denote vectors (1\myst order tensors) as small bold letters \(\myvector{v}\), matrices (2\mynd order tensors) as bold capital letters \(\mymatrix{M}\) and tensors, which generalize the concept of matrices for orders (number of dimensions) higher than 2, in capital calligraphic letters  \(\mytensor{X}\). The \emph{n--mode} product is defined, for a given tensor \(\mytensor{X} \in \myR^{D_0 \times D_1 \times \cdots \times D_N}\) and a matrix \( \mymatrix{M} \in \myR^{R \times D_n} \), as the tensor \(\mytensor{T} = \mytensor{X} \times_n \mymatrix{M} \in \myR^{D_0 \times \cdots \times D_{n-1} \times R \times D_{n+1} \times \cdots \times D_N} \), with:
%\begin{equation}\nonumber
\(
   \mytensor{T}_{i_0, i_1, \cdots, i_n} = \sum_{k=0}^{D_n} \mymatrix{M}_{i_n, k} \mytensor{X}_{i_0, i_1, \cdots, i_n} 
\).
%\end{equation}

\textbf{Latent high-order parametrization of the network:}
We introduce tensor factorization in the context of deep neural networks. Note that this method is independent of the dimensionality of the input but we introduce it here, without loss of generality, for the case of a \(4\) dimensional kernel of 2D convolutions. Specifically, we consider a deep neural network composed of \(L\) layers convolutional layers, interlaced with non-linearities \(\Phi_l\), \(l \in \myrange{1}{L}\). 
Let's consider a convolutional layer \(l \in \myrange{1}{L}\), taking as input an activation tensor \(\mytensor{X}_l\) and parametrized by a weight tensor \(\mytensor{W}_l \in \myR^{F, C, H, W}\), where  \(F, C, H, W\) correspond respectively to number of Filters (e.g. output channels), input Channels, Height and Width of the convolutional kernel. The output of that layer, after applying non-linearity, will be \(\Phi \left( \mytensor{X}_l \myconv \mytensor{W}_l \right)\).

We introduce a latent parametrization of the weight kernel \(\mytensor{W}_l\) by expressing it as a low-rank tensor, in this paper using a Tucker decomposition~\cite{kolda2009tensor}. In other words we express \(\mytensor{W}_l\) in a latent subspace as a core tensor \(\mytensor{G}_l\). The mapping to and from this subspace is done via factor matrices \(\mymatrix{U}_l^{F}, \mymatrix{U}_l^{C},\mymatrix{U}_l^{H} \text{ and } \mymatrix{U}_l^{W}\):
%\begin{equation}
%    \label{eq:low-rank-weight}
\(
    \mytensor{W}_l = \mytensor{G}_l \times_0 \mymatrix{U}_l^{F} \times_1 \mymatrix{U}_l^{C} \times_2 \mymatrix{U}_l^{H} \times_3 \mymatrix{U}_l^{W}
\).
%\end{equation}

\textbf{Tensor dropout: } 
In addition to the above deterministic decomposition, we apply tensor dropout~\cite{kolbeinsson2021tensor} to each layer, effectively randomizing the rank of the decomposition. To do so, we introduce diagonal sketching matrices \(\mymatrix{M}_{F}, \mymatrix{M}_{C},\mymatrix{M}_{H} \text{ and } \mymatrix{M}_{W}\), the diagonal entries of which are i.i.d. and follow a Bernoulli distribution parametrized by probability \(\theta \in [0, 1]\).
Specifically, we samples random vectors \(\myvector{\lambda}^{F}  \in \myR^O, \myvector{\lambda}^{C}  \in \myR^C, \myvector{\lambda}^{H}  \in \myR^H \text{ and } \myvector{\lambda}^{W}  \in \myR^W\), the entries of which are i.i.d. following a Bernoulli distribution parametrized by probability \(\theta\).  We can then define the sketching matrices as \( \mymatrix{M}_{O} = \mydiag(\myvector{\lambda}_{F}), \mymatrix{M}_{C} = \mydiag(\myvector{\lambda}_{C}), \mymatrix{M}_{H} = \mydiag(\myvector{\lambda}_{H}) \text{ and } \mymatrix{M}_{W} = \mydiag(\myvector{\lambda}_{W}) \). 

This randomization is then applied not directly to the weight tensor \(\mytensor{W}\), but rather in the low-rank subspace, effectively randomizing the \emph{rank} of the convolutional kernel:
\begin{equation}\label{eq:bernouilli-tucker-weight}
    \mytensor{\tilde W}_l 
    = 
    \underbrace{
        \left( \mytensor{G}_l \times_0  \mymatrix{M}_F \times \cdots \times_{3} \mymatrix{M}_W\right)
    }_{\text{randomized core } \mytensor{\hat G}_l}
    \times_0 
    \underbrace{
        \left(\mymatrix{U}^{F}_l\mymatrix{M}_F\myT \right)
        \times
        \cdots
        \times_{3} \left(\mymatrix{U}^{W}_l\mymatrix{M}_W\myT \right)
    }_{\text{randomized factors } \mymatrix{\hat U}^{F}_l,  
                                  \mymatrix{\hat U}^{I}_l,  
                                  \mymatrix{\hat U}^{H}_l,
                                  \mymatrix{\hat U}^{W}_l}
\end{equation}

This stochastic reduction of the rank can be done without affecting performance thanks to the over-parametrization of deep networks, which, while crucial for learning~\citep{du2018power, soltanolkotabi2018theoretical}, create large amounts of redundancies.
In addition, since \(\mytensor{G}_l \times_0  \mymatrix{M}_F \times_0 \left(\mymatrix{U}^{F}_l \mymatrix{M}_F\myT \right) = \mytensor{G}_l \times_0 \left(\mymatrix{U}^{F}_l\mymatrix{M}_F\myT \mymatrix{M}_F \right)\), and \(\mymatrix{M}_F, \mymatrix{M}_I, \mymatrix{M}_H \text{ and } \mymatrix{M}_W\) are idempotent, eq.~\ref{eq:bernouilli-tucker-weight} can be simplified to:
\begin{equation}\label{eq:bernouilli-tucker-weight-fast}
    \mytensor{\tilde W}_l 
    = 
    \left( \mytensor{G}_l \times_0  \mymatrix{M}_F \times \cdots \times_{3} \mymatrix{M}_W\right)
    \times_0 
        \mymatrix{U}^{F}_l
        \times
        \cdots
        \times_{3} \mymatrix{U}^{W}_l
    =
    \mytensor{\tilde G}_l \times_0 
        \mymatrix{U}^{F}_l
        \times
        \cdots
        \times_{3} \mymatrix{U}^{W}_l
\end{equation}
In other words, we sketch the core tensor, then project it back using the original factors.
Importantly, the randomization terms from the above equation, \(\mytensor{\tilde G} = \mytensor{G} \times_0  \mymatrix{M}_{F} \times \cdots \times_{N} \mymatrix{M}_{W}\) are never explicitly computed using actual tensor contractions. Instead, the elements are directly sampled from the core and the corresponding factors, which is much more computationally effective. For the binary case, we plug in eq.~(\ref{eq:bernouilli-tucker-weight-fast}) into the binarization procedure.

The randomization being done in the latent subspace, it induces no sparsity, unlike pruning or dropout based methods and the reconstructed weights are dense. Since the weights are learnt end-to-end with randomization on the latent cores, the network cannot rely on any single latent component for prediction, thus learning intrinsically more robust representations. The result is a network that is naturally more robust to perturbations in the inputs.

\begin{figure*}[htb]
  \centering
  \includegraphics[width=0.9\textwidth]{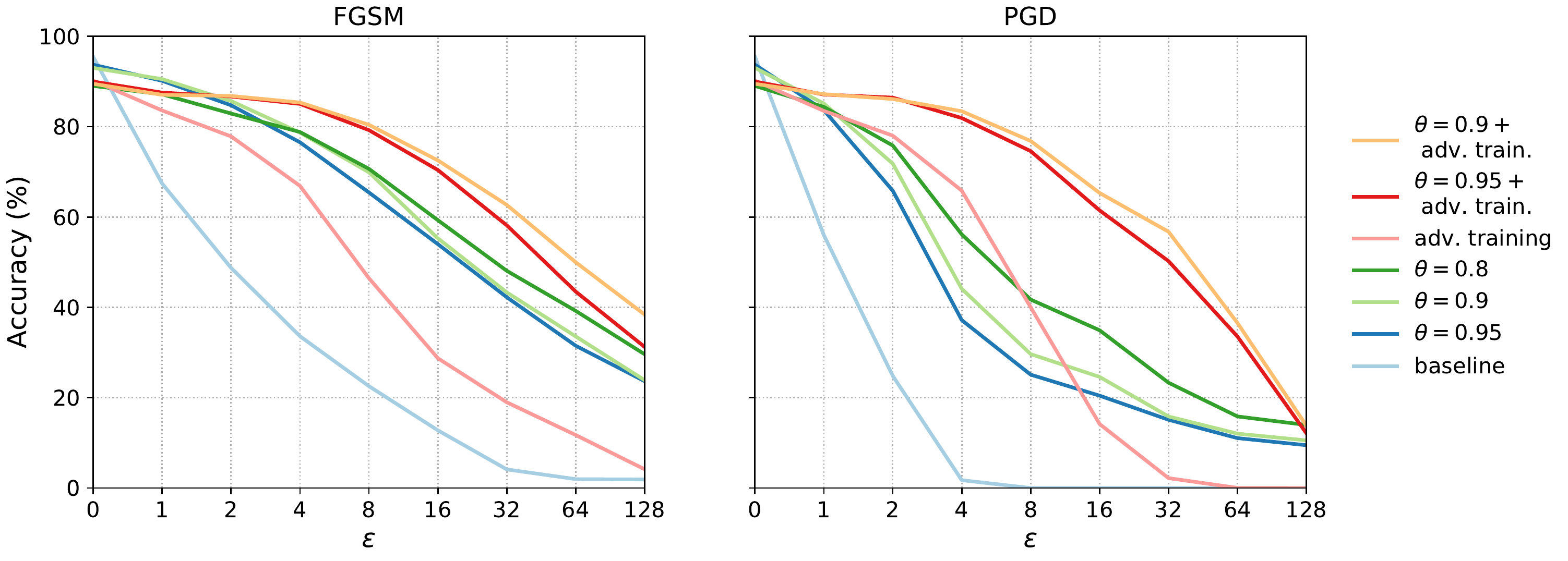}
  \caption{FGSM~(left)
  and PGD~(right)
  attacks on the CIFAR-10 dataset for various values of $\epsilon$ with and without adversarial training. Notice that our method alone surpasses the strong adversarial training defence. When combined together the robustness is increased even further.}
  \label{fig:cifar-fgsm}
\end{figure*}

\section{Experimental setting and implementation details}\label{sec:experimental-setup}

\textbf{Datasets:}
We conducted experiments on two widely used datasets for image and audio classification, CIFAR-10 and Speech Command~\citep{speechcomand2}.
\textbf{CIFAR-10~\citep{krizhevsky2009learning}} is a widely used image classification dataset consisting of \(60,000\) images ($50,000$ training and $10,0000$ testing) of size \(32 \times 32\)px in \(10\) classes, equally represented. We only used random horizontal flipping for data augmentation.
See supplementary material for details on Speech Command.

\textbf{Training the model}
All our CIFAR-10~\citep{krizhevsky2009learning} experiments were conducted using a ResNet-18~\citep{he2016deep} architecture. The network was trained for $350$ epochs using SGD with momentum ($0.9$) and a starting learning rate of $0.1$ that was dropped at epoch $150$ and $250$ by a factor of $0.1$. The weight decay was set to $1e-6$. In order to accelerate the training process the models with $\theta<1$ were initialized from a pretrained model that was trained without stochasticity ($\theta=1$). The binary counterpart models were trained following the method proposed by~\citet{rastegari2016xnor} using the same optimizer and learning scheduler as for the real-valued ones. 
The binarization and re-parametrization follows the same procedure as for CIFAR-10. See supplementary material for details on the speech recognition experiments. Training was done on a single 1080Ti GPU.
\begin{wraptable}[17]{r}{0.49\textwidth}
    \small
    % \vspace{-0.5cm}
    \resizebox{0.49\textwidth}{!}{
    \begin{tabular}{cccccc}
    \toprule[1.2pt]
    \multirow{2}{*}{\textbf{Attack}} & \multirow{2}{*}{\(\epsilon\)}      &  \multirow{2}{*}{\textbf{Baseline}} & \multicolumn{3}{c}{\textbf{Tensor Dropout Rate $\theta$}}\\
    \cline{4-6}
    \addlinespace[0.1cm]
    &  &  &  \(\mathbf{0.95}\) &  \(\mathbf{0.9}\) &  \(\mathbf{0.8}\) \\
    \toprule[1pt]
    \multicolumn{2}{c}{\textbf{Clean}} & \multirow{2}{*}{$\mathbf{95.3}$} & \multirow{2}{*}{$94.5$} & \multirow{2}{*}{ $93.0$ } & \multirow{2}{*}{ $90.1$} \\
    \multicolumn{2}{c}{\emph{(no attack)}} &  &  & &  \\
    \midrule[0.8pt]
    \parbox[t]{2mm}{\multirow{3}{*}{\rotatebox[origin=c]{90}{\textbf{FGSM}}}}
        & \textbf{2 } &  $48.7$ & $84.9$ & $\mathbf{86.8}$ & $83.4$  \\
        & \textbf{8 } &  $22.5$ & $65.4$ & $69.9$ & $\mathbf{71.5}$  \\
        & \textbf{16} &  $12.7$ & $54.0$ & $56.0$ & $\mathbf{60.3}$  \\
    \midrule[0.8pt]
    \parbox[t]{2mm}{\multirow{3}{*}{\rotatebox[origin=c]{90}{\textbf{BIM}}}}
        & \textbf{2 } &  $23.0$ & $60.2$ & $69.5$ & $\mathbf{71.8}$  \\
        & \textbf{8 } &  $0.0$ & $26.6$ & $33.1$ & $\mathbf{45.5}$  \\
        & \textbf{16} &  $0.0$ & $27.0$ & $33.0$ & $\mathbf{42.4}$  \\
        \midrule[0.8pt]
    \parbox[t]{2mm}{\multirow{3}{*}{\rotatebox[origin=c]{90}{\textbf{PGD}}}}
        & \textbf{2 } &  $22.9$ & $64.4$ & $72.3$ & $\mathbf{76.2}$  \\
        & \textbf{8 } &  $0.0$ & $27.0$ & $28.1$ & $\mathbf{42.9}$  \\
        & \textbf{16} &  $0.0$ & $22.4$ & $27.4$ & $\mathbf{34.3}$  \\
    \bottomrule[1.2pt]
    \end{tabular}
    }
    \caption{\small\textbf{Real-valued network performance on CIFAR-10} for FGSM, BIM and PGD attacks.}\label{tab:main-results-real}
\end{wraptable}

\textbf{Attacking the model:} 
For FGSM, we run the attack for various values of $\epsilon = \{1, 2, 4, 8, 16,\allowbreak 32, 64, 128\}$ (for an image range \myrange{0}{255}) across the entire validation/testing set averaging the results over 10 runs. 
For the iterative methods BIM and PGD, we follow~\citet{kurakin2016adversarial} and~\citet{song2017pixeldefend} setting the step size to $1$ and the number of iterations to $\floor{\text{min}(\epsilon+4,1.25\epsilon)}$. 

\textbf{Threat Model: } we assume that the attacker has access to everything (e.g. the architecture of the network, its weights, inputs, outputs, training process and gradients, etc) except the random seed used for sampling the Bernoulli random variables.

All of our models were implemented using PyTorch~\citep{paszke2017automatic} and trained on a single Titan X GPU. The latent, randomized tensor factorization was implemented using TensorLy~\citep{kossaifi2019tensorly}. For the adversarial attacks we used the FoolBox 2.0~\citep{rauber2017foolbox} and AutoAttack~\citep{croce2020reliable} packages.

\begin{table}
    \small
    %\vspace{-0.5cm}
    %\resizebox{0.49\textwidth}{!}{
    \begin{tabular}{ccccccccc}
    \toprule[1.2pt]
    \multirow{2}{*}{\textbf{Attack}} & \multirow{2}{*}{\(\epsilon\)}      &  \multirow{2}{*}{\textbf{Baseline}} & \multicolumn{6}{c}{\textbf{Tensor Dropout Rate $\theta$}}\\
    \cline{4-9}
    \addlinespace[0.1cm]
    &  &  &  \(\mathbf{0.95}\) &  \(\mathbf{0.9}\) &  \(\mathbf{0.8}\) & \(\mathbf{0.95}\) + adv. &  \(\mathbf{0.9}\) + adv. &  \(\mathbf{0.8}\) + adv. \\
    \toprule[1pt]
    \multicolumn{2}{c}{\textbf{Clean}} & \multirow{2}{*}{$\mathbf{76.2}$} & \multirow{2}{*}{$68.9$} & \multirow{2}{*}{ $67.2$ } & \multirow{2}{*}{ $62.6$} & \multirow{2}{*}{ $60.4$} & \multirow{2}{*}{ $58.6$} & \multirow{2}{*}{ $56.7$} \\
    \multicolumn{2}{c}{\emph{(no attack)}} &  &  & & & & &  \\
    \midrule[0.8pt]
    \parbox[t]{2mm}{\multirow{3}{*}{\rotatebox[origin=c]{90}{\textbf{FGSM}}}}
        & \textbf{2 } &  $21.8$ & $65.8$ & $64.5$ & $58.8$ & $58.5$ & $58.5$ & $54.8$  \\
        & \textbf{8 } &  $8.7$ & $57.6$ & $56.5$ & $54.2$ & $57.1$ & $56.2$ & $52.6$ \\
        & \textbf{16} &  $4.5$ & $47.9$ & $48.9$ & $48.3$ & $53.2$ & $54.2$ & $51.3$   \\
    \midrule[0.8pt]
    \parbox[t]{2mm}{\multirow{3}{*}{\rotatebox[origin=c]{90}{\textbf{BIM}}}}
        & \textbf{2 } &  $11.0$ & $56.6$ & $55.7$ & $54.4$ & $56.9$ & $58.0$ & $53.9$  \\
        & \textbf{8 } &  $0.0$ & $40.6$ & $44.1$ & $44.7$ & $53.4$ & $53.6$ & $51.2$ \\
        & \textbf{16} &  $0.0$ & $37.6$ & $41.8$ & $43.0$ & $50.2$ & $52.6$ & $50.6$  \\
        \midrule[0.8pt]
    \parbox[t]{2mm}{\multirow{3}{*}{\rotatebox[origin=c]{90}{\textbf{PGD}}}}
        & \textbf{2 } &  $11.6$ & $57.5$ & $58.3$ & $55.3$ & $56.8$ & $57.4$ & $55.2$  \\
        & \textbf{8 } &  $0.0$ & $36.7$ & $38.6$ & $41.8$ & $54.5$ & $53.8$ & $51.8$  \\
        & \textbf{16} &  $0.0$ & $26.6$ & $29.0$ & $32.0$ & $50.0$ & $49.8$ & $49.5$ \\
    \bottomrule[1.2pt]
    \end{tabular}
    %}
    \caption{\small\textbf{Real-valued network performance on CIFAR-100} for FGSM, BIM and PGD attacks. + adv indicated training with both $\theta$ and adversarial training.}\label{tab:main-results-real-cifar100}
\end{table}

\section{Results}
In this section, we empirically demonstrate the robustness property of our proposed method against adversarial examples by extensively evaluating it on CIFAR-10 and comparing it against existing state of the art defense techniques.  Moreover, we show that our approach can be combined with adversarial training based techniques leading to further robustness gains. All the experiments are run using the experimental setup described in Section~\ref{sec:experimental-setup}.

\begin{wraptable}[8]{r}{0.49\textwidth}
    \small
    % \vspace{-0.5cm}
    \resizebox{0.49\textwidth}{!}{
    \begin{tabular}{cccccc}
    \toprule[1.2pt]
    \multirow{2}{*}{\textbf{Attack}} & \multirow{2}{*}{\(\epsilon\)}      &  \multirow{2}{*}{\textbf{Baseline}} & \multicolumn{3}{c}{\textbf{Tensor Dropout Rate $\theta$}}\\
    \cline{4-6}
    \addlinespace[0.1cm]
    &  &  &  \(\mathbf{0.95}\) &  \(\mathbf{0.9}\) &  \(\mathbf{0.8}\) \\
    \toprule[1pt]
    AA     & \textbf{8 } &  $0.0$ & $30.0$ &  $35.4$ & $37.5$  \\
    \bottomrule[1.2pt]
    \end{tabular}
    }
    \caption{\small\textbf{Real-valued network performance on CIFAR-10} using AutoAttack~\cite{croce2020reliable}}\label{tab:main-results-real-aa}
\end{wraptable}
\textbf{Robustness to adversarial attacks: }
When evaluated against FGSM attacks on CIFAR-10, our method is significantly more robust than the baseline approach, especially for high values of $\epsilon$ (see Table~\ref{tab:main-results-real} and Fig.~\ref{fig:cifar-fgsm}).  Furthermore, the results presented in Table~\ref{tab:main-results-real} show that for lower values of $\theta=0.8$, our network significantly decreases the typical high attack success-rate achieved by strong iterative attack algorithms such as BIM and PGD. We note however that against newer attack sets, such AutoAttack~\citep{croce2020reliable}, our model is noticeable less robust. We leave further exploration of this area within the tensors framework for future work.

\textbf{Comparison to the State of the Art: } To better understand the performance of our method, we compare it against existing state of theart defense techniques such as: DQ~\citep{lin2019defensive} in which the authors attempt to reduce the propagation of the adversarial attacks inside the network by controlling the Lipschitz constant, features squeezing~\citep{xu2017feature} that simple reduces the dimensionality of the search space by controlling the color bit depth of each pixel while applying spatial smoothing and finally, adversarial training in two variants: R+FGSM as proposed in~\citep{lin2019defensive} and PGD~\citep{madry2017towards}. Where the later (i.e. adversarial training) is considered to be one of the strongest defence techniques developed. For feature-squeezing we used 5 bits for image color reduction combined with a $2\times2$ median filter. For adversarial training as in~\citep{madry2017towards,kurakin2016adversarial} we sample the number of steps (for PGD) and $\epsilon$ randomly. As the results from Table~\ref{tab:comparison-defense-sota} show, our method consistently outperforms existing defense strategies for various attacks (FGSM, BIM, PGD), across different values of $\epsilon=\{2,8,16\}$. Furthermore, when combined with adversarial training our method can further increase its resilience to attacks.
\begin{table}[ht]
    \centering
    \small
    \begin{tabular}{cccc}
        \toprule
        Method & clean & \begin{tabular}{@{}c@{}}\textbf{FGSM} \\ $\epsilon=2/8/16$\end{tabular} & \begin{tabular}{@{}c@{}}\textbf{PGD} \\ $\epsilon=2/8/16$\end{tabular} \\
        \midrule
        Normal & $95.4$ & $48.7/22.6/12.9$ & $22.9/0/0$ \\
        DQ~\citep{lin2019defensive} & $\mathbf{95.9}$ & $68/53/42$ & $62/4/0$ \\
        Feature squeezing~\citep{xu2017feature} & $94.1$ & $61/35/27$ & $64/2/0$ \\
        Adv. training+R FGSM~\citep{lin2019defensive}& $91.6$ & $81/52/38$ & $84/43/11$ \\
        Adv. training PGD~\citep{madry2017towards}& $86.6$ & $74/46/31$ & $76/44/20$ \\
        \textbf{Ours ($\theta=0.8$)} & $90.1$ & $83.4/71.5/60.3$ & $76.2/42.9/34.3$ \\
        \textbf{Ours ($\theta=0.95$) + Adv. training PGD} & $89.5$ & $\mathbf{86.5}/\mathbf{81.4}/\mathbf{70.1}$ & $\mathbf{84.9}/\mathbf{75.4}/\mathbf{59.8}$ \\
        \bottomrule
    \end{tabular}
        \caption{\small \textbf{Comparison against various defense methods}
    % against FGSM and PGD} with $\epsilon=\{2,8,16\}$ 
    on CIFAR10. Our method significantly outperforms other state of theart methods, especially when combined with adversarial training.}
    
    \label{tab:comparison-defense-sota}
\end{table}

\textbf{Defending against black-box attacks}
While we focus mostly on white box attacks, we also show that our approach increases robustness against black-box attacks, following the same standard setting as~\citet{mustafa2019adversarial}. Our method is more robust in all cases Table~\ref{table:black-box}.

\begin{wraptable}[7]{r}{0.6\textwidth}
\centering
\resizebox{0.6\textwidth}{!}{
    \begin{tabular}{lllll}\toprule Method & Clean & \begin{tabular}{@{}c@{}}\textbf{FGSM} \\ $\epsilon=2/8/16$\end{tabular} & \begin{tabular}{@{}c@{}}\textbf{BIM} \\ $\epsilon=2/8/16$\end{tabular} & \begin{tabular}{@{}c@{}}\textbf{PGD} \\ $\epsilon=2/8/16$\end{tabular} \\ \midrule Baseline & 95.4 & 94.2/87.8/79.1 & 93.0/84.7/77.5 & 94.0/82.3/59.9 \\ \midrule Ours & 89.5 & 87.4/87.1/83.3 & 87.4/84.5/83.8 & 87.6/85.2/83.4 \\ \bottomrule  \end{tabular}
}
\caption{\textbf{Defending against Black-box attacks}~\cite{mustafa2019adversarial}}\label{table:black-box}
\end{wraptable}
One thing to notice is the relative difference in performance: our method starts with a slightly lower performance on the clean set but is much less affected by the adversarial attacks.

\textbf{Defending against omniscient attackers: } Another interesting question is whether our defense strategy would still work against an \emph{omniscient} attacker, i.e., an attacker with access to the full un-randomized weights. We trained a network for \(\theta = 0.9\). Then, during training, we first generate an adversarial example using the full, un-randomized weights (i.e. \(\theta = 1\)) and test it using the stochasticity (i.e. \(\theta = 0.9\)). As can be seen in Table~\ref{tab:robustness-un-randomized}, our network is still robust against these attacks, despite the attacker having full access to the weights. Similar behaviour can be observed for other values of $\theta$. 
 Note that this is an extreme scenario and in general, the weights could be stored separately (and safely) with the network getting, at each time, the randomly reconstructed weights.

\textbf{Defending against adaptive attacks:}
Given that the proposed method hardens the models against adversarial attacks via a randomisation performed in a latent subspace, herein we analyse its resilience against adaptive methods typically capable of overcoming randomization-based approaches. In particular, following~\cite{athalye2018obfuscated} at each iteration of gradient descent, for each convolutional layer, instead of taking a step in the direction of $\nabla_x f(x)$  we move in the direction of $\sum_{i=1}^{k}\nabla_x f(x)$ where each pass has the weights randomized in the latent space using our approach. While in~\cite{athalye2018obfuscated} the authors use $k=10$ we try with up to $k=20$ but without noticing any significant increase in the success rate of the attack. The PGD attack itself was run for 500 iterations as in~\cite{athalye2018obfuscated}. The experiment reported was conducted on CIFAR 10 using our best model. As the results from Table~\ref{tab:defense-bpda} show, our method is capable of offering a reasonable degree of robustness even against such attack that probe wherever the defense technique mask or shatter the gradients.

\begin{table}
    \centering
    \begin{subtable}{.63\textwidth}
    \caption{\textbf{Robustness of our method against BPDA on CIFAR10}.}
    \label{tab:defense-bpda}
    \small
    \centering
    \begin{tabular}{cccc}
        \toprule
       Attack / $\eps$ & 2 & 8 & 16 \\
        \midrule
        BPDA~\citep{athalye2018obfuscated} & $83.3$ & $54.9$ & $43.8$ \\
        \bottomrule
    \end{tabular}
    \caption{\textbf{Robustness against attacks on CIFAR10 for methods applying the randomization on the subspace (ours) vs directly on the activations}. The network used and attack settings are aligned with the ones used in~\citet{wang2018defensive}.}
    \label{tab:latent-comparison}
    \begin{tabular}{ccccc}
        \toprule
        Method & \textbf{Clean} & \textbf{FGSM} & \textbf{BIM} & \textbf{PGD}\\
        \midrule
        Ours & $85.9$ & $60.0$ & $42.9$ & $43.8$ \\
        ~\cite{wang2018defensive} & $83.4$ & $41.7$ & $32.3$ & $35.2$ \\
        \bottomrule
    \end{tabular}
    \end{subtable}
        \hfill
    \begin{subtable}{.35\textwidth}
        \caption{\textbf{Robustness against attacker with access to the un-randomized weights}. FGSM, BIM and PGD attacks with~$\epsilon \in \{2,8,16\}$ are computed using the full (un-randomized) weights (i.e. \(\theta=1\)) and used against the \textbf{same} network with the \textbf{same} weights but with \(\theta = 0.9\).}
    \label{tab:robustness-un-randomized}
    \small
    \begin{tabular}{cccc}
        \toprule
        $\epsilon$ & \textbf{FGSM} & \textbf{PGD} & \textbf{BIM}\\
        \midrule
        \textbf{2} & $91.1$ & $82.5$ & $81.9$ \\
        \textbf{8} & $85.4$ & $74.1$ & $80.4$ \\
        \textbf{16} & $82.4$ & $53.9$ & $80.1$ \\
        \bottomrule
    \end{tabular}
    \end{subtable}
    
    \caption{Robustness in various scenarios on CIFAR10.}
    \label{tab:grouped_tables}
    \vspace{-15px}
\end{table}

\section{Ablation studies}
To further validate our findings we test out approach on two different scenarios: on fully binarized networks~(Section~\ref{ssec:attack-on-binary}) and audio classification task~(see supplementary material). Furthermore, we empirically validate the importance of performing the randomization in a latent subspace~(Section~\ref{ssec:latent-comparison}) and that of using a tensorized form of the algorithm over its matrixised version~(Section~\ref{ssec:matrix-vs-tensor}). See supplementary material for results on speech recognition.

\textbf{Importance of randomization in a latent subspace }\label{ssec:latent-comparison}
Here, we demonstrate the importance of incorporating the stochasticity in the latent subspace rather than  in the parameter space directly. 
%To further showcase the advantages of our approach against ones that don’t make the randomization in the subspace 
We compare our method against the so-called \textit{Defensive dropout} of~\cite{wang2018defensive}, in which the authors proposed to apply dropout directly to the activation of the first fully connected layer at test time. Since their method requires the presence of multiple fully connected layers, we apply our randomized tensorization directly on their architecture. The results of this comparison can be seen in Table~\ref{tab:latent-comparison} in the same setting and the same epsilon as in the original paper~\cite{wang2018defensive}. Notice that our approach consistently outperforms the defensive dropout.

\textbf{Matrix vs. Tensor decomposition}\label{ssec:matrix-vs-tensor}
% Here, we empirically validate the advantage of our tensor-based method over the more traditional matrix approach.% for our particular setting.
Tensor methods have the ability to leverage the multi-linear structure in the data, weight and activations. This structure is typically discarded by matrix based approaches, thus loosing information that we would want to preserve and leverage when building robust neural networks. 

%\begin{table}[h!]
\begin{wraptable}[8]{r}{0.6\textwidth}
% \vspace{-0.45cm}
\centering
    \resizebox{0.6\textwidth}{!}{
    \begin{tabular}{ccccc}
        \toprule
        Method & \textbf{Clean} &   \begin{tabular}{@{}c@{}}\textbf{FGSM} \\ (2/8/16)\end{tabular} & \begin{tabular}{@{}c@{}}\textbf{BIM} \\ (2/8/16)\end{tabular} & \begin{tabular}{@{}c@{}}\textbf{PGD} \\ (2/8/16)\end{tabular}\\
        \midrule
        Ours & $94.5$ & $84.9/65.4/54.0$ & $60.6/26.6/27.0$ & $64.4/27.0/22.4$ \\
        Matrix & $93.7$ & $76.9/52.8/40.9$ & $44.6/17.5/18.2$ & $50.0/16.9/15.0$ \\
        \bottomrule
    \end{tabular}
    }
    \caption{\textbf{Robustness against attacks on CIFAR10 for the tensor decomposition (ours) and matrix case}.}
    \label{tab:matrix-vs-tensor}
\end{wraptable}
It can be noted that the matrix case is a special case of our approach. Specifically, in Equation~\ref{eq:bernouilli-tucker-weight}, this can be obtained by setting $M_I = M_H = M_W = U_l^{I} = U_l^{H}  = U_l^{W} = \mathbf{I}$. The equality can then be rewritten in term of the mode-1 unfolding to obtain the matrix case. 
To demonstrate the advantage of our tensor-based approach compared to the matrix case, we ran an additional experiment on CIFAR-10, using a ResNet-18 architecture for both the matrix and tensor version of our method, for the same value $\theta$. As the results from Table~\ref{tab:matrix-vs-tensor} show, the tensor decompositions offers consistent gains over the matrix one.

\textbf{Attacking binary neural networks.}\label{ssec:attack-on-binary} %\vspace{-0.2cm}
Network binarization is the most extreme case of quantization, where the weights and features are represented using a single bit~\citep{courbariaux2016binarized, rastegari2016xnor, bulat2017binarized}. The typical approach quantizes the network using the \textit{sign} function~\citep{courbariaux2015binaryconnect}, however this introduces high quantization errors that hinder the learning process. To alleviate this, a real-valued scaling factor is introduced by~\citet{rastegari2016xnor}. In this work we binarize the network following~\citet{rastegari2016xnor}:
%\begin{equation}\label{eq:binarization}
    $\mytensor{I} \ast \mytensor{W} = \left(\text{sgn}(\mytensor{I}) \oast \text{sgn}(\mytensor{W}) \right) \odot \mytensor{K} \myvector{\alpha}$,
%\end{equation}
where $\mytensor{I}\in\mathbb{R}^{c \times w_{in} \times h_{in}}$ and $\mytensor{W}\in \mathbb{R}^{c \times h \times w}$ denote the input and the weight of the $L$-th convolutional layer, $\myvector{\alpha} \in \mathbb{R}^{c \times 1 \times 1}$ represent the weight scaling factor and $\mytensor{K}\in\mathbb{R}^{1 \times h_{out} \times w_{out}}$ the input scaling factor. Both $\myvector{\alpha}$ and $\mytensor{K}$ are computed analytically as proposed by~\citet{rastegari2016xnor}.
\begin{wraptable}[19]{r}{0.49\textwidth}
    \large
    % \vspace{-0.5cm}
    \resizebox{0.49\textwidth}{!}{
    \begin{tabular}{cccccc}
    \toprule[1.1pt]
    \multicolumn{2}{l}{\multirow{2}{*}{\textbf{Attack} \, $\epsilon$}} 
    % & \multirow{2}{*}{\(\epsilon\)}     
    &  \multicolumn{3}{c}{\textbf{Baseline}} & \textbf{Ours}\\
    \cline{3-5}
    \addlinespace[0.1cm]
    &  &  Id &  tanh($x$) &  tanh(0.75$x$) &  $\theta$=0.99 \\
    \toprule[1pt]
    \multicolumn{2}{c}{\textbf{Clean}} & \multirow{2}{*}{$83.7$} & \multirow{2}{*}{$83.7$} & \multirow{2}{*}{ $83.7$ } & \multirow{2}{*}{ $80.0$} \\
    \multicolumn{2}{c}{\emph{(no attack)}} &  &  & &  \\
    \midrule[0.7pt]
    \parbox[t]{2mm}{\multirow{3}{*}{\rotatebox[origin=c]{90}{\textbf{FGSM}}}}
        & \textbf{2 } &  $36.6$ & $34.5$ & $34.1$ & $\mathbf{76.9}$ \\
        & \textbf{8 } &  $6.9$ & $6.1$ &  $5.8$ & $\mathbf{65.0}$  \\
        & \textbf{16} &  $4.3$ & $3.4$ & $3.0$ & $\mathbf{58.7}$   \\
    \midrule[0.7pt]
    \parbox[t]{2mm}{\multirow{3}{*}{\rotatebox[origin=c]{90}{\textbf{BIM}}}}
        & \textbf{2 } &  $37.0$ & $34.7$  & $35.1$ & $\mathbf{66.3}$ \\
        & \textbf{8 } &  $0.0$ & $0.0$ & $0.0$ & $\mathbf{46.4}$   \\
        & \textbf{16} &  $0.0$ & $0.0$ & $0.0$ & $\mathbf{44.0}$ \\
        \midrule[0.7pt]
    \parbox[t]{2mm}{\multirow{3}{*}{\rotatebox[origin=c]{90}{\textbf{PGD}}}}
        & \textbf{2 } &  $41.7$  & $38.7$ & $39.8$ & $\mathbf{67.5}$  \\
        & \textbf{8 } &  $0.1$ & $0.0$ & $0.0$ & $\mathbf{47.9}$   \\
        & \textbf{16} &  $0.0$ & $0.0$ & $0.0$ & $\mathbf{41.5}$   \\
    \bottomrule[1.1pt]
    \end{tabular}
    }
    \caption{\small \textbf{Binary network performance on CIFAR-10} for FGSM, BIM and PGD attacks. Our approach is significantly more robust, especially against iterative attacks.}
    \label{tab:main-results-binary}
    \vspace{0.3cm}
\end{wraptable}
While it was previously thought that such binarized networks are more resilient to adversarial attacks~\citep{khalil2018combinatorial,liu2018adv,galloway2017attacking} than their real-valued counterpart, in this work we confirm the recent findings of~\citep{lin2019defensive} by showing that in fact it is the opposite, i.e., binary networks are more susceptible to adversarial attacks. Typically, during the training phase the derivative of the quantization function (sgn) is approximated using a STE (e.g., an identity function clipped to $[-1,1]$ in this case). The same estimator can be used during the attacking phase and it often leads to a high rate of success of the attacks. Interestingly, the results in Table~\ref{tab:main-results-binary} show that we can go one step further by approximating the derivative of the $\textit{sgn}$ function using tanh($x$) and tanh($0.75x$) respectively. The use of these approximations make the binary networks become more sensitive to the attacks. 
When evaluated on a binary network on the CIFAR-10 dataset, as for the real-valued models, our method shows significant improvements across the entire range of values and attacks tested (see Table~\ref{tab:main-results-binary} and Fig.~\ref{fig:cifar-fgsm}). We note that for the binary case, since such networks have a lower representational capacity, we set $\theta=0.99$.

\section{Conclusion}\label{sec:conclusion}
Defensive tensorization is a novel adversarial defence technique that leverages a latent high order factorization of the network. Tensor dropout is applied in the latent subspace, directly on the factorized weights parametrizing each layer, resulting in dense reconstructed weights, without the sparsity or perturbations typically induced by randomization. We empirically demonstrate that this approach makes the network significantly more robust to adversarial attacks. Contrarily to a widely spread belief, we observe that binary networks are \emph{more} sensitive to adversarial attacks than their real-valued counter-part. 
We show that our method significantly improves robustness in the face of adversarial attacks for both binary and real-valued networks. We demonstrate this empirically for both image and audio classification.

\appendix

\section{Effect on the optimization landscape} 
\begin{figure}[ht!]
  \centering
  \includegraphics[width=\linewidth]{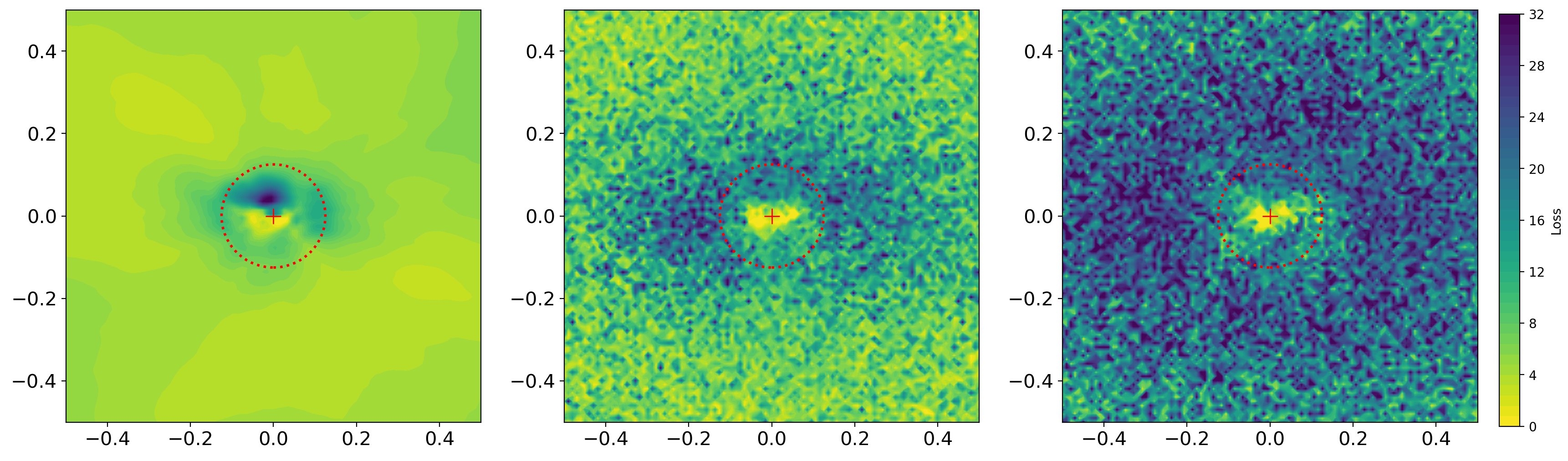}
\caption{\textbf{Contour plot of the loss surface} of our model, adversarially trained model for various values of $\theta$ evaluated on the {\em l}$_\infty$ neighbourhood of an unseen CIFAR-10 image. The same direction was used for all three plots. The red circle denotes the $\epsilon=32$ neighbourhood. }
  \label{fig:loss-adv-real}
\end{figure}

To visually assess the impact of both our randomization scheme 
%and adversarial training 
on the optimization landscape, we visualise the evaluation of the loss in a fixed neighbourhood around an unseen data point. Specifically, we visualize the loss function learned by the model in the neighbourhood of new, unseen data point $x$.
For clarity, we visualise the loss in a 2--dimensional space, along direction of the gradient at $x$ (x-axis) and a randomly chosen direction, orthogonal to the direction of gradient (y-axis). Next, a mesh-grid is constructed by sampling uniformly points along these two directions for the range $[-0.5, 0.5]$. Then a contour plot is constructed by evaluating the losses for all points on the mesh-grid. The result for our best model can be seen in figure~\ref{fig:loss-adv-real}.

Intuitively, the randomization (which is done in the \emph{latent} subspace of the decomposition, not on the weights themselves), changes the loss function, at each pass, making it hard to converge to a fixed attack due to the presence of many spurious minimums. This can be seen by looking at the landscape of the loss function around an arbitrary sample in Fig.~\ref{fig:loss-adv-real}. The landscape is inline with the finding of~\citet{madry2017towards}, where the authors show the adversarial training smooths the loss space around 0. This is even more noticeable for the method that combines our approach with adversarial training (see Fig. ~\ref{fig:loss-adv-real}).

\section{Defensive tensorization for audio classification}\label{ssec:audio-results}

To further demonstrate the generalizability of our approach, this section considers adversarial attacks on the audio domain, measuring the efficacy of our method.

\subsection{Experimental setting and implementation details}

\noindent\textbf{Speech Command:} Speech Command~\citep{speechcomand2} is an audio recognition dataset comprised of $105,000$ 1-second utterances of words from a large number of users spanning over a small vocabulary. The objective is to recognize among ten spoken words: {\em yes, no, up, down, left, right, on, off, stop, go}, in addition to recognizing words outside the vocabulary as {\em unknown}, and detecting {\em silence}. The dataset is balanced and all audio recordings are captured with a sampling frequency of 16 KHz. We use a 80\%-10\%-10\% splits for training, validation and testing respectively.

\noindent\textbf{Implementation details:} For the experiments conducted on the Speech Command dataset we build on the SoundNet5~\citep{aytar2016soundnet} architecture containing 5 convolutional layers \emph{[in\_channels, out\_channels, kernel, stride, padding]}: $[1,16,(1\times64),(1\times2),(0\times32)], [16,32,(1\times32),(1\times2),(0\times16)], [32,64,(1\times16),(1\times2),(0\times8)], [64,128,(1\times8),(1\times2),(0\times4)],
[128,256,(1\times4),(1\times2),(0\times2)]$ and 2 linear ones:  $[512, 256]$ and $[256,12]$. Each convolutional layer was followed by a max-pooling operation. We trained all of the audio models using Adam~\citep{kingma2014adam} for $50$ epochs with an initial learning rate set to $0.01$ that was dropped by $0.1\times$ at epoch $25$ and $35$.

\noindent\textbf{Attacking the model:} The attack model largely follows the procedure used for images, with a small adaptation: On the Speech Command dataset, since the raw data is in the range $[-1,1]$, we scaled the value of $\epsilon$ accordingly, running it for the following values $\epsilon=\{0.008, 0.032, 0.063\}$.

\subsection{Results}

 In-line with the latest success in audio recognition, we consider an end-to-end audio model following SoundNet \citep{aytar2016soundnet} architecture, that operates directly on the raw audio signal, without requiring any feature extractions (e.g., MFCC or log mel-spectrogram). We found that the end-to-end models show higher degree of vulnerability to the adversarial attacks, e.g., around 6\% absolute degradation compared to the model operating on log mel-spectrogram. In case of the small vocabulary audio recognition task, we only consider FGSM attack and summarize our findings in Table~\ref{tab:main-results-real-audio}. With a higher degree of stochasticity (i.e. $\theta=0.9$), both the real and the binarized model exhibit much higher resilience to the adversarial attacks.   
\begin{table}
    \centering
    % \small
    \caption{\textbf{Performance on Speech Command} for FGSM attacks using both binary and real-valued models. Notice that our approach is significantly more robust.}
    \label{tab:main-results-real-audio}
    % \resizebox{0.49\textwidth}{!}{
    \begin{tabular}{cccccc}
    \toprule[1.2pt]
    \multirow{2}{*}{\textbf{Quant.}} & \multirow{2}{*}{\(\epsilon\)}      &  \multirow{2}{*}{\textbf{Baseline}} & \multicolumn{3}{c}{\textbf{Defensive Tensorization}}\\
    \addlinespace[3pt]
    \cline{4-6}
    \addlinespace[3pt]
    &  &  &  \(\mathbf{\theta=0.99}\) &  \(\mathbf{\theta=0.95}\) &  \(\mathbf{\theta=0.9}\) \\
    \toprule[1pt]
    \parbox[t]{2mm}{\multirow{4}{*}{\rotatebox[origin=c]{90}{\textbf{Real}}}}
        & \textbf{No attack} &  $\mathbf{93.8}$ & $92.0$ & $89.6$ & $88.1$  \\
        \addlinespace[2pt]
        \cline{2-6}
        \addlinespace[2pt]
        & \textbf{0.008} &  $33.0$ & $49.6$ & $58.2$ & $\mathbf{61.0}$  \\
        & \textbf{0.032} &  $14.9$ & $33.0$ & $40.2$ & $\mathbf{44.2}$  \\
        & \textbf{0.063} &  $7.6$ & $23.8$ & $31.8$ & $\mathbf{35.7}$  \\
    \midrule[0.8pt]
    \parbox[t]{2mm}{\multirow{4}{*}{\rotatebox[origin=c]{90}{\textbf{Binary}}}}
      & \textbf{No attack} &  $88.0$ & $\mathbf{89.0}$ & $83.5$ & $83.2$  \\
      \addlinespace[2pt]
      \cline{2-6}
      \addlinespace[2pt]
         & \textbf{0.008} &  $12.2$ & $50.1$ & $54.4$ & $\mathbf{56.0}$ \\
         & \textbf{0.032} &  $3.0$ & $31.5$ & $35.6$ & $\mathbf{40.2}$  \\
         & \textbf{0.063} &  $0.2$ & $26.7$ & $30.3$ & $\mathbf{30.5}$  \\
    \bottomrule[1.2pt]
    \end{tabular}
    % }
\end{table}

\bibliography{references}

\begin{thebibliography}{57}
\providecommand{\natexlab}[1]{#1}
\providecommand{\url}[1]{\texttt{#1}}
\expandafter\ifx\csname urlstyle\endcsname\relax
  \providecommand{\doi}[1]{doi: #1}\else
  \providecommand{\doi}{doi: \begingroup \urlstyle{rm}\Url}\fi

\bibitem[Akhtar and Mian(2018)]{akhtar2018threat}
Naveed Akhtar and Ajmal Mian.
\newblock Threat of adversarial attacks on deep learning in computer vision: A
  survey.
\newblock \emph{IEEE Access}, 6:\penalty0 14410--14430, 2018.

\bibitem[Amodei et~al.(2016)Amodei, Olah, Steinhardt, Christiano, Schulman, and
  Man{\'e}]{amodei2016concrete}
Dario Amodei, Chris Olah, Jacob Steinhardt, Paul Christiano, John Schulman, and
  Dan Man{\'e}.
\newblock Concrete problems in ai safety.
\newblock \emph{arXiv preprint arXiv:1606.06565}, 2016.

\bibitem[Astrid and Lee(2017)]{astrid2017cp}
Marcella Astrid and Seung{-}Ik Lee.
\newblock Cp-decomposition with tensor power method for convolutional neural
  networks compression.
\newblock \emph{CoRR}, abs/1701.07148, 2017.

\bibitem[Athalye et~al.(2018)Athalye, Carlini, and
  Wagner]{athalye2018obfuscated}
Anish Athalye, Nicholas Carlini, and David Wagner.
\newblock Obfuscated gradients give a false sense of security: Circumventing
  defenses to adversarial examples.
\newblock \emph{arXiv preprint arXiv:1802.00420}, 2018.

\bibitem[Battaglino et~al.(2018)Battaglino, Ballard, and
  Kolda]{battaglino2018practical}
Casey Battaglino, Grey Ballard, and Tamara~G Kolda.
\newblock A practical randomized cp tensor decomposition.
\newblock \emph{SIAM Journal on Matrix Analysis and Applications}, 39\penalty0
  (2):\penalty0 876--901, 2018.

\bibitem[Bulat and Tzimiropoulos(2017)]{bulat2017binarized}
Adrian Bulat and Georgios Tzimiropoulos.
\newblock Binarized convolutional landmark localizers for human pose estimation
  and face alignment with limited resources.
\newblock In \emph{ICCV}, 2017.

\bibitem[Carlini and Wagner(2017)]{carlini2017towards}
Nicholas Carlini and David Wagner.
\newblock Towards evaluating the robustness of neural networks.
\newblock In \emph{2017 IEEE Symposium on Security and Privacy (SP)}, pages
  39--57. IEEE, 2017.

\bibitem[Cohen et~al.(2019)Cohen, Rosenfeld, and Kolter]{cohen2019certified}
Jeremy Cohen, Elan Rosenfeld, and Zico Kolter.
\newblock Certified adversarial robustness via randomized smoothing.
\newblock In \emph{International Conference on Machine Learning}, pages
  1310--1320. PMLR, 2019.

\bibitem[Courbariaux et~al.(2015)Courbariaux, Bengio, and
  David]{courbariaux2015binaryconnect}
Matthieu Courbariaux, Yoshua Bengio, and Jean-Pierre David.
\newblock Binaryconnect: Training deep neural networks with binary weights
  during propagations.
\newblock In \emph{NIPS}, 2015.

\bibitem[Courbariaux et~al.(2016)Courbariaux, Hubara, Soudry, El-Yaniv, and
  Bengio]{courbariaux2016binarized}
Matthieu Courbariaux, Itay Hubara, Daniel Soudry, Ran El-Yaniv, and Yoshua
  Bengio.
\newblock Binarized neural networks: Training deep neural networks with weights
  and activations constrained to+ 1 or-1.
\newblock \emph{arXiv}, 2016.

\bibitem[Croce and Hein(2020)]{croce2020reliable}
Francesco Croce and Matthias Hein.
\newblock Reliable evaluation of adversarial robustness with an ensemble of
  diverse parameter-free attacks.
\newblock In \emph{International conference on machine learning}, pages
  2206--2216. PMLR, 2020.

\bibitem[Dhillon et~al.(2018)Dhillon, Azizzadenesheli, Lipton, Bernstein,
  Kossaifi, Khanna, and Anandkumar]{dhillon2018stochastic}
Guneet~S Dhillon, Kamyar Azizzadenesheli, Zachary~C Lipton, Jeremy Bernstein,
  Jean Kossaifi, Aran Khanna, and Anima Anandkumar.
\newblock Stochastic activation pruning for robust adversarial defense.
\newblock \emph{arXiv preprint arXiv:1803.01442}, 2018.

\bibitem[Dong et~al.(2018)Dong, Liao, Pang, Su, Zhu, Hu, and
  Li]{dong2018boosting}
Yinpeng Dong, Fangzhou Liao, Tianyu Pang, Hang Su, Jun Zhu, Xiaolin Hu, and
  Jianguo Li.
\newblock Boosting adversarial attacks with momentum.
\newblock In \emph{Computer Vision and Pattern Recognition}, 2018.

\bibitem[Du and Lee(2018)]{du2018power}
Simon~S Du and Jason~D Lee.
\newblock On the power of over-parametrization in neural networks with
  quadratic activation.
\newblock In \emph{ICML}, 2018.

\bibitem[Erichson et~al.(2017)Erichson, Manohar, Brunton, and
  Kutz]{erichson2017randomized}
N~Benjamin Erichson, Krithika Manohar, Steven~L Brunton, and J~Nathan Kutz.
\newblock Randomized cp tensor decomposition.
\newblock \emph{arXiv preprint arXiv:1703.09074}, 2017.

\bibitem[Galloway et~al.(2017)Galloway, Taylor, and
  Moussa]{galloway2017attacking}
Angus Galloway, Graham~W Taylor, and Medhat Moussa.
\newblock Attacking binarized neural networks.
\newblock \emph{arXiv preprint arXiv:1711.00449}, 2017.

\bibitem[Goodfellow et~al.(2014)Goodfellow, Shlens, and
  Szegedy]{goodfellow2014explaining}
Ian~J Goodfellow, Jonathon Shlens, and Christian Szegedy.
\newblock Explaining and harnessing adversarial examples.
\newblock \emph{arXiv preprint arXiv:1412.6572}, 2014.

\bibitem[Gowal et~al.(2019)Gowal, Dvijotham, Stanforth, Bunel, Qin, Uesato,
  Arandjelovic, Mann, and Kohli]{gowal2019scalable}
Sven Gowal, Krishnamurthy~Dj Dvijotham, Robert Stanforth, Rudy Bunel, Chongli
  Qin, Jonathan Uesato, Relja Arandjelovic, Timothy Mann, and Pushmeet Kohli.
\newblock Scalable verified training for provably robust image classification.
\newblock In \emph{Proceedings of the IEEE/CVF International Conference on
  Computer Vision}, pages 4842--4851, 2019.

\bibitem[Gowal et~al.(2020)Gowal, Qin, Uesato, Mann, and
  Kohli]{gowal2020uncovering}
Sven Gowal, Chongli Qin, Jonathan Uesato, Timothy Mann, and Pushmeet Kohli.
\newblock Uncovering the limits of adversarial training against norm-bounded
  adversarial examples.
\newblock \emph{arXiv preprint arXiv:2010.03593}, 2020.

\bibitem[Guo et~al.(2017)Guo, Rana, Cisse, and van~der
  Maaten]{guo2017countering}
Chuan Guo, Mayank Rana, Moustapha Cisse, and Laurens van~der Maaten.
\newblock Countering adversarial images using input transformations.
\newblock \emph{arXiv preprint arXiv:1711.00117}, 2017.

\bibitem[He et~al.(2016)He, Zhang, Ren, and Sun]{he2016deep}
Kaiming He, Xiangyu Zhang, Shaoqing Ren, and Jian Sun.
\newblock Deep residual learning for image recognition.
\newblock In \emph{Computer Vision and Pattern Recognition}, 2016.

\bibitem[Huang et~al.(2017)Huang, Papernot, Goodfellow, Duan, and
  Abbeel]{huang2017adversarial}
Sandy Huang, Nicolas Papernot, Ian Goodfellow, Yan Duan, and Pieter Abbeel.
\newblock Adversarial attacks on neural network policies.
\newblock \emph{arXiv preprint arXiv:1702.02284}, 2017.

\bibitem[Khalil et~al.(2018)Khalil, Gupta, and
  Dilkina]{khalil2018combinatorial}
Elias~B Khalil, Amrita Gupta, and Bistra Dilkina.
\newblock Combinatorial attacks on binarized neural networks.
\newblock \emph{arXiv preprint arXiv:1810.03538}, 2018.

\bibitem[Kim et~al.(2016)Kim, Park, Yoo, Choi, Yang, and
  Shin]{yong2016compression}
Yong{-}Deok Kim, Eunhyeok Park, Sungjoo Yoo, Taelim Choi, Lu~Yang, and Dongjun
  Shin.
\newblock Compression of deep convolutional neural networks for fast and low
  power mobile applications.
\newblock \emph{ICLR}, 05 2016.

\bibitem[Kolbeinsson et~al.(2021)Kolbeinsson, Kossaifi, Panagakis, Bulat,
  Anandkumar, Tzoulaki, and Matthews]{kolbeinsson2021tensor}
Arinbj{\"o}rn Kolbeinsson, Jean Kossaifi, Yannis Panagakis, Adrian Bulat,
  Animashree Anandkumar, Ioanna Tzoulaki, and Paul~M Matthews.
\newblock Tensor dropout for robust learning.
\newblock \emph{IEEE Journal of Selected Topics in Signal Processing},
  15\penalty0 (3):\penalty0 630--640, 2021.

\bibitem[Kolda and Bader(2009)]{kolda2009tensor}
Tamara~G. Kolda and Brett~W. Bader.
\newblock Tensor decompositions and applications.
\newblock \emph{SIAM REVIEW}, 51\penalty0 (3):\penalty0 455--500, 2009.

\bibitem[Kossaifi et~al.(2019{\natexlab{a}})Kossaifi, Bulat, Tzimiropoulos, and
  Pantic]{kossaifi2019t}
Jean Kossaifi, Adrian Bulat, Georgios Tzimiropoulos, and Maja Pantic.
\newblock T-net: Parametrizing fully convolutional nets with a single
  high-order tensor.
\newblock In \emph{Computer Vision and Pattern Recognition},
  2019{\natexlab{a}}.

\bibitem[Kossaifi et~al.(2019{\natexlab{b}})Kossaifi, Panagakis, Anandkumar,
  and Pantic]{kossaifi2019tensorly}
Jean Kossaifi, Yannis Panagakis, Anima Anandkumar, and Maja Pantic.
\newblock Tensorly: Tensor learning in python.
\newblock \emph{The Journal of Machine Learning Research}, 20\penalty0
  (1):\penalty0 925--930, 2019{\natexlab{b}}.

\bibitem[Krizhevsky and Hinton(2009)]{krizhevsky2009learning}
Alex Krizhevsky and Geoffrey Hinton.
\newblock Learning multiple layers of features from tiny images.
\newblock Technical report, Citeseer, 2009.

\bibitem[Kurakin et~al.(2016)Kurakin, Goodfellow, and
  Bengio]{kurakin2016adversarial}
Alexey Kurakin, Ian Goodfellow, and Samy Bengio.
\newblock Adversarial examples in the physical world.
\newblock \emph{arXiv preprint arXiv:1607.02533}, 2016.

\bibitem[Lebedev et~al.(2015)Lebedev, Ganin, Rakhuba, Oseledets, and
  Lempitsky]{lebedev2015speeding}
Vadim Lebedev, Yaroslav Ganin, Maksim Rakhuba, Ivan~V. Oseledets, and Victor~S.
  Lempitsky.
\newblock Speeding-up convolutional neural networks using fine-tuned
  cp-decomposition.
\newblock In \emph{ICLR}, 2015.

\bibitem[Lin et~al.(2019)Lin, Gan, and Han]{lin2019defensive}
Ji~Lin, Chuang Gan, and Song Han.
\newblock Defensive quantization: When efficiency meets robustness.
\newblock \emph{arXiv preprint arXiv:1904.08444}, 2019.

\bibitem[Liu et~al.(2018)Liu, Li, Wu, and Hsieh]{liu2018adv}
Xuanqing Liu, Yao Li, Chongruo Wu, and Cho-Jui Hsieh.
\newblock Adv-bnn: Improved adversarial defense through robust bayesian neural
  network.
\newblock \emph{arXiv preprint arXiv:1810.01279}, 2018.

\bibitem[Madry et~al.(2017)Madry, Makelov, Schmidt, Tsipras, and
  Vladu]{madry2017towards}
Aleksander Madry, Aleksandar Makelov, Ludwig Schmidt, Dimitris Tsipras, and
  Adrian Vladu.
\newblock Towards deep learning models resistant to adversarial attacks.
\newblock \emph{arXiv preprint arXiv:1706.06083}, 2017.

\bibitem[Mustafa et~al.(2019)Mustafa, Khan, Hayat, Goecke, Shen, and
  Shao]{mustafa2019adversarial}
Aamir Mustafa, Salman Khan, Munawar Hayat, Roland Goecke, Jianbing Shen, and
  Ling Shao.
\newblock Adversarial defense by restricting the hidden space of deep neural
  networks.
\newblock In \emph{International Conference on Computer Vision}, 2019.

\bibitem[Novikov et~al.(2015)Novikov, Podoprikhin, Osokin, and
  Vetrov]{novikov2015tensorizing}
Alexander Novikov, Dmitry Podoprikhin, Anton Osokin, and Dmitry Vetrov.
\newblock Tensorizing neural networks.
\newblock In \emph{Neural Information Processing Systems}, 2015.

\bibitem[Paszke et~al.(2017)Paszke, Gross, Chintala, Chanan, Yang, DeVito, Lin,
  Desmaison, Antiga, and Lerer]{paszke2017automatic}
Adam Paszke, Sam Gross, Soumith Chintala, Gregory Chanan, Edward Yang, Zachary
  DeVito, Zeming Lin, Alban Desmaison, Luca Antiga, and Adam Lerer.
\newblock Automatic differentiation in {PyTorch}.
\newblock In \emph{NIPS Autodiff Workshop}, 2017.

\bibitem[Rastegari et~al.(2016)Rastegari, Ordonez, Redmon, and
  Farhadi]{rastegari2016xnor}
Mohammad Rastegari, Vicente Ordonez, Joseph Redmon, and Ali Farhadi.
\newblock Xnor-net: Imagenet classification using binary convolutional neural
  networks.
\newblock In \emph{ECCV}, 2016.

\bibitem[Rauber et~al.(2017)Rauber, Brendel, and Bethge]{rauber2017foolbox}
Jonas Rauber, Wieland Brendel, and Matthias Bethge.
\newblock Foolbox v0. 8.0: A python toolbox to benchmark the robustness of
  machine learning models.
\newblock \emph{arXiv preprint arXiv:1707.04131}, 5, 2017.

\bibitem[Rice et~al.(2020)Rice, Wong, and Kolter]{rice2020overfitting}
Leslie Rice, Eric Wong, and Zico Kolter.
\newblock Overfitting in adversarially robust deep learning.
\newblock In \emph{International Conference on Machine Learning}, pages
  8093--8104. PMLR, 2020.

\bibitem[Samangouei et~al.(2018)Samangouei, Kabkab, and
  Chellappa]{samangouei2018defense}
Pouya Samangouei, Maya Kabkab, and Rama Chellappa.
\newblock Defense-gan: Protecting classifiers against adversarial attacks using
  generative models.
\newblock \emph{arXiv preprint arXiv:1805.06605}, 2018.

\bibitem[Sidiropoulos et~al.(2014)Sidiropoulos, Papalexakis, and
  Faloutsos]{sidiropoulos2014parallel}
Nicholas~D Sidiropoulos, Evangelos~E Papalexakis, and Christos Faloutsos.
\newblock Parallel randomly compressed cubes: A scalable distributed
  architecture for big tensor decomposition.
\newblock \emph{IEEE Signal Processing Magazine}, 31\penalty0 (5):\penalty0
  57--70, 2014.

\bibitem[Soltanolkotabi et~al.(2018)Soltanolkotabi, Javanmard, and
  Lee]{soltanolkotabi2018theoretical}
Mahdi Soltanolkotabi, Adel Javanmard, and Jason~D Lee.
\newblock Theoretical insights into the optimization landscape of
  over-parameterized shallow neural networks.
\newblock \emph{IEEE Transactions on Information Theory}, 2018.

\bibitem[Song et~al.(2017)Song, Kim, Nowozin, Ermon, and
  Kushman]{song2017pixeldefend}
Yang Song, Taesup Kim, Sebastian Nowozin, Stefano Ermon, and Nate Kushman.
\newblock Pixeldefend: Leveraging generative models to understand and defend
  against adversarial examples.
\newblock \emph{arXiv preprint arXiv:1710.10766}, 2017.

\bibitem[Tsourakakis(2010)]{tsourakakis2010mach}
Charalampos~E Tsourakakis.
\newblock Mach: Fast randomized tensor decompositions.
\newblock In \emph{Proceedings of the 2010 SIAM International Conference on
  Data Mining}, pages 689--700. SIAM, 2010.

\bibitem[Vervliet et~al.(2014)Vervliet, Debals, Sorber, and
  De~Lathauwer]{vervliet2014breaking}
Nico Vervliet, Otto Debals, Laurent Sorber, and Lieven De~Lathauwer.
\newblock Breaking the curse of dimensionality using decompositions of
  incomplete tensors: Tensor-based scientific computing in big data analysis.
\newblock \emph{IEEE Signal Processing Magazine}, 31\penalty0 (5):\penalty0
  71--79, 2014.

\bibitem[Wang et~al.(2018)Wang, Wang, Zhao, Wen, Kaeli, Chin, and
  Lin]{wang2018defensive}
Siyue Wang, Xiao Wang, Pu~Zhao, Wujie Wen, David Kaeli, Peter Chin, and Xue
  Lin.
\newblock Defensive dropout for hardening deep neural networks under
  adversarial attacks.
\newblock In \emph{Proceedings of the International Conference on
  Computer-Aided Design}, page~71. ACM, 2018.

\bibitem[Wang et~al.(2015)Wang, Tung, Smola, and Anandkumar]{wang2015fast}
Yining Wang, Hsiao-Yu Tung, Alexander~J Smola, and Anima Anandkumar.
\newblock Fast and guaranteed tensor decomposition via sketching.
\newblock In \emph{Neural Information Processing Systems}, 2015.

\bibitem[Wang et~al.(2020)Wang, Zou, Yi, Bailey, Ma, and Gu]{wang2020improving}
Yisen Wang, Difan Zou, Jinfeng Yi, James Bailey, Xingjun Ma, and Quanquan Gu.
\newblock Improving adversarial robustness requires revisiting misclassified
  examples.
\newblock In \emph{International Conference on Learning Representations}, 2020.

\bibitem[{Warden}(2018)]{speechcomand2}
Pete {Warden}.
\newblock {Speech Commands: A Dataset for Limited-Vocabulary Speech
  Recognition}.
\newblock \emph{arXiv e-prints}, art. arXiv:1804.03209, Apr 2018.

\bibitem[Xie et~al.(2018)Xie, Wang, Zhang, Ren, and Yuille]{xie2018mitigating}
Cihang Xie, Jianyu Wang, Zhishuai Zhang, Zhou Ren, and Alan Yuille.
\newblock Mitigating adversarial effects through randomization.
\newblock In \emph{International Conference on Learning Representations}, 2018.

\bibitem[Xie et~al.(2019)Xie, Wu, Maaten, Yuille, and He]{xie2019feature}
Cihang Xie, Yuxin Wu, Laurens van~der Maaten, Alan~L Yuille, and Kaiming He.
\newblock Feature denoising for improving adversarial robustness.
\newblock In \emph{Computer Vision and Pattern Recognition}, 2019.

\bibitem[Xu et~al.(2017)Xu, Evans, and Qi]{xu2017feature}
Weilin Xu, David Evans, and Yanjun Qi.
\newblock Feature squeezing: Detecting adversarial examples in deep neural
  networks.
\newblock \emph{arXiv preprint arXiv:1704.01155}, 2017.

\bibitem[Yang et~al.(2019)Yang, Zhang, Katabi, and Xu]{yang2019me}
Yuzhe Yang, Guo Zhang, Dina Katabi, and Zhi Xu.
\newblock Me-net: Towards effective adversarial robustness with matrix
  estimation.
\newblock \emph{ICML}, 2019.

\bibitem[Yuan et~al.(2019)Yuan, Li, Cao, and Zhao]{yuan2019randomized}
Longhao Yuan, Chao Li, Jianting Cao, and Qibin Zhao.
\newblock Randomized tensor ring decomposition and its application to
  large-scale data reconstruction.
\newblock \emph{arXiv preprint arXiv:1901.01652}, 2019.

\bibitem[Zhang et~al.(2019)Zhang, Chen, Xiao, Gowal, Stanforth, Li, Boning, and
  Hsieh]{zhang2019towards}
Huan Zhang, Hongge Chen, Chaowei Xiao, Sven Gowal, Robert Stanforth, Bo~Li,
  Duane Boning, and Cho-Jui Hsieh.
\newblock Towards stable and efficient training of verifiably robust neural
  networks.
\newblock \emph{arXiv preprint arXiv:1906.06316}, 2019.

\bibitem[Zhou et~al.(2014)Zhou, Cichocki, and Xie]{zhou2014decomposition}
Guoxu Zhou, Andrzej Cichocki, and Shengli Xie.
\newblock Decomposition of big tensors with low multilinear rank.
\newblock \emph{arXiv preprint arXiv:1412.1885}, 2014.

\end{thebibliography}

\end{document}